\pdfoutput=1
\documentclass[letterpaper, 10 pt, conference]{ieeeconf}  %

\IEEEoverridecommandlockouts                              %

\usepackage{amsmath, amsfonts, amssymb}
\usepackage{graphicx}
\usepackage{xcolor}
\usepackage{float}
\usepackage{cite}
\usepackage{bm}

\usepackage{algorithm}
\usepackage{algpseudocode}

\usepackage{booktabs}
\usepackage{multirow}

\usepackage{hyperref}

\newcommand{\mat}[1]{\bm{#1}}
\renewcommand{\vec}[1]{\bm{#1}}

\newcommand{\zeros}{\mathbf{0}}

\newcommand{\st}{\vec{s}} %
\newcommand{\ot}{\vec{o}} %
\newcommand{\at}{\vec{a}} %

\renewcommand{\v}{\vec{v}}

\graphicspath{{figures/}}

\algrenewcommand\algorithmicrequire{\textbf{Input:}}
\algrenewcommand\algorithmicensure{\textbf{Output:}}

\usepackage{mathtools}
\usepackage{array}

\newif\ifanonymous
\anonymousfalse         %

\title{\LARGE \bf
PushMimic: Learning Adaptive Humanoid Box Pushing via \\
Dynamic Motion Augmentation and Reference Distillation
}

\ifanonymous
\author{Anonymous Authors}
\else
\author{
Yulun Zhuang, Hanyu Zhang, Jingyu Liu and Yanran Ding$^{*}$%
\thanks{
$^*$ Corresponding author email: {\tt\small yanrand@umich.edu}.
Authors are with the Department of Robotics, University of Michigan, Ann Arbor, MI 48109, USA.
}%
}
\fi

\begin{document}

\maketitle
\thispagestyle{empty}
\pagestyle{empty}

\begin{abstract}
Humanoid box pushing is challenging when the physical properties of the task, such as the box mass and the ramp inclination, are unknown before the push begins.
This paper presents PushMimic, a framework that learns adaptive humanoid box pushing from a single level-ground kinematic reference motion.
The learned policy adapts its pushing posture to these properties without observing them.
PushMimic augments the reference motion into a dynamically feasible motion dataset over these properties with sampling-based trajectory optimization, and distills the dataset into one motion tracking policy in one stage of training.
Simulation results over box masses from $5$ to $15~\mathrm{kg}$ and ramp inclinations up to $15^\circ$ show that PushMimic raises the success rate from $57.8\%$ to $88.8\%$ compared with a baseline that tracks the original reference motion.
It also reduces the cost of push of the arms by $22\%$.
Hardware experiments on a Unitree G1 humanoid show that the policy adapts its torso pitch during continuous pushing on an inclined treadmill, with box masses up to $19~\mathrm{kg}$ and ramp inclinations up to $12.6^\circ$.
PushMimic therefore needs only one reference motion, instead of one for every box mass and ramp inclination.
\end{abstract}

\section{Introduction}\label{sec:intro}

Humanoid robots are increasingly used in repetitive and physically demanding loco-manipulation tasks in manufacturing and logistics, such as sorting parcels and relocating boxes~\cite{yu2026omnicontact, chen2026scenebot, hu2026vofa}, mostly in structured factory settings. However, pushing a large box is still challenging, since the physical properties of the task, such as the box mass, the friction, and the ground inclination, are unknown to the robot before the push begins and vary from one task to another. For example, pushing a box up a ramp requires the robot to lean its torso further forward than on level ground. A human adjusts this posture through the interaction with the box after the push begins. A humanoid robot that keeps the same posture for every task consumes more energy and can fail to deliver the box accurately. Therefore, the humanoid robot should also adapt its posture during the push through the interaction with the box.

Existing solutions for humanoid box pushing are often based on reinforcement learning (RL), either from task rewards~\cite{zhang2025falcon, hu2026vofa, li2025amo, xu2025facet} or by tracking human reference motions~\cite{cheng2024expressive, fu2024humanplus, liao2026beyondmimic, weng2025hdmi, zhao2025resmimic} retargeted from motion capture datasets~\cite{mahmood2019amass, harvey2020robust, li2023object}.
A task reward alone allows many feasible ways to push a box, such as different gaits and different body parts in contact with the box.
Choosing one of them through reward design is not intuitive and requires much trial and error.
A human reference motion specifies the gait and the contacts together, so tracking it is a practical alternative~\cite{zhang2026learning}.
However, the available human pushing motions are few and recorded on level ground~\cite{li2023object, yu2026omnicontact}.
Recording one motion for every combination of the physical properties of the task, denoted as task parameters, is impractical.
Therefore, a practical setting is to start from one level-ground pushing motion, denoted as the nominal reference motion, and derive the reference motions for the other task parameters from it.
The problem is then to learn, from the nominal reference motion, one policy that pushes the box over a range of task parameters and adapts its posture without observing them.

\begin{figure}[t]
    \centering
    \includegraphics[width=\columnwidth]{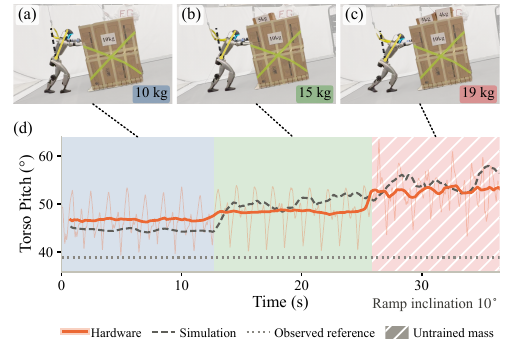}
    \caption{Online posture adaptation of the G1 when weights are added to the box during continuous pushing on a treadmill inclined at $10^\circ$. (a) to (c) Snapshots with box masses of $10$, $15$, and $19~\mathrm{kg}$, where the angle between the belt normal (dashed) and the line from the pelvis to the top of the head (yellow) illustrates the torso pitch. (d) Torso pitch from the belt normal on hardware and in simulation, averaged over one stride, and that of the nominal reference motion that the policy observes. The faint curve is the raw measurement. The background colors mark the box mass of each snapshot, and the hatching marks the masses beyond the $15~\mathrm{kg}$ upper bound of the training grid.}
    \label{fig:hw_mass}
    \vspace{-1em}
\end{figure}

Policies trained from task rewards alone can adapt the pushing force to the task.
FALCON~\cite{zhang2025falcon} applies increasing external forces at the end effectors during training to learn a force-adaptive whole-body controller.
VOFA~\cite{hu2026vofa} builds on FALCON and combines the force-adaptive controller with a visual goal policy to push objects of unknown mass and friction.
FACET~\cite{xu2025facet} tracks a virtual impedance reference, and force control policies~\cite{zhi2025learning, portela2024learning} estimate the external force from the state history without a force sensor.
However, in these methods, the posture comes from the reward rather than from human motion.
The same holds when the pushing behavior comes from a planning cost~\cite{zhang2026sumo} or when the policy imitates only part of a reference motion~\cite{das2026partial}.
Choosing the pushing posture therefore remains a reward or cost design problem.
Policies that track a human reference motion take the posture from the reference instead.
This approach has enabled humanoids to carry, push, and stack boxes and to interact with other objects~\cite{chen2026scenebot, yu2026omnicontact, weng2025hdmi, zhao2025resmimic, yang2025omniretarget}.
However, the reference motion is recorded under one task parameter.
Other task parameters are usually covered by domain randomization during training, so the policy keeps the recorded posture for all of them.
Therefore, neither family provides a posture that both follows human motion and adapts to the task parameter.

\begin{figure*}
    \centering
    \includegraphics[width=\textwidth]{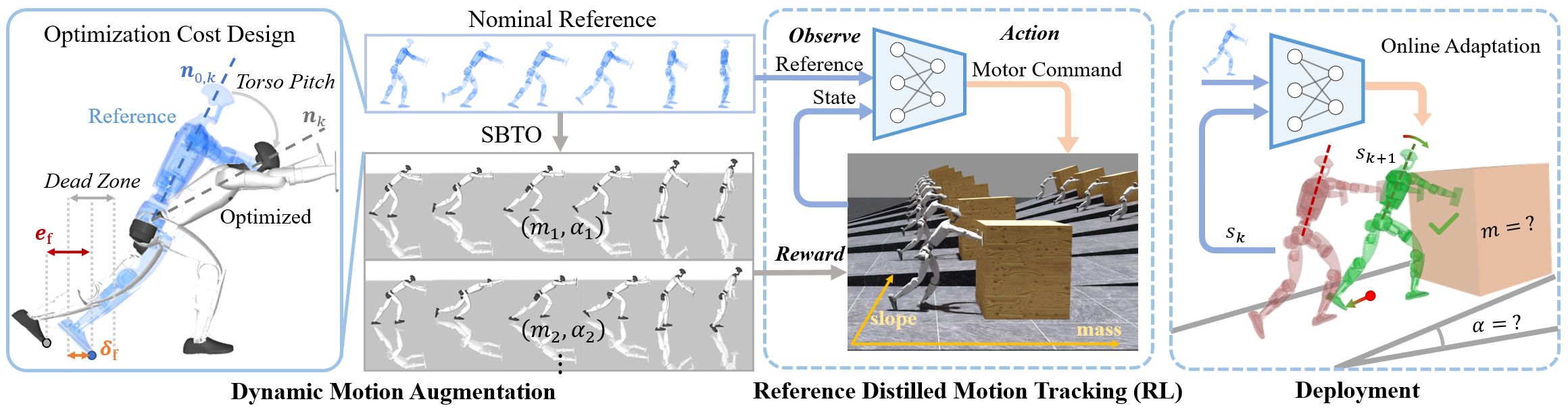}
    \caption{Overview of PushMimic. \textbf{Dynamic motion augmentation}: sampling-based trajectory optimization (SBTO)~\cite{dhedin2026dynaretarget} turns the nominal reference motion into one dynamically feasible reference motion for each box mass and ramp inclination on a grid, using a torso pitch prior and a foot placement dead zone. \textbf{Reference distilled motion tracking}: an RL policy observes only the nominal reference motion and the states of the robot and the box, and is rewarded for tracking the augmented reference motion of its task parameter. \textbf{Deployment}: the policy adapts its posture to the box mass and the ramp inclination without observing them.}
    \label{fig:teaser}
    \vspace{-1em}
\end{figure*}

A reference motion for a new task parameter does not have to be recorded, since recent works generate new reference motions from existing ones.
Motion generators sample them from a learned motion distribution~\cite{rempe2026kimodo, zhang2026learning}.
OmniRetarget~\cite{yang2025omniretarget} re-solves the kinematic retargeting of one reference motion for new object and terrain configurations.
AdaMimic~\cite{huang2025towards} edits the keyframes of one reference motion for a new task parameter and leaves the rest to a motion tracking policy.
However, these methods work at the kinematic level, and the dynamics is only considered when the motion tracking policy is trained.
In box pushing, the task parameters change the posture through the contact forces.
For example, a steeper ramp makes the torso lean further forward and moves the feet further back.
Kinematic methods do not model these forces, so they cannot predict how the posture should change with the task parameters.
Therefore, the reference motions for new task parameters should be generated with the dynamics of the robot and the box.

Dynamically feasible retargeting can generate such reference motions.
It optimizes a control sequence that tracks a kinematic reference motion in a physics simulator, so the resulting motion accounts for the contact forces.
SPIDER~\cite{pan2025spider} guides the sampling with virtual contacts, DynaRetarget~\cite{dhedin2026dynaretarget} optimizes the full motion through expanding horizons, and contact-implicit shooting~\cite{esteban2026shooting} embeds a differentiable simulation in a nonlinear program.
These methods are mostly used to make a reference motion dynamically feasible under the task parameter at which it was recorded.
DynaRetarget has also augmented one reference motion for objects of different masses and shapes.
Therefore, running the same optimization under a new task parameter gives a dynamically feasible motion that stays close to the nominal reference motion.
Such a motion can serve as the reference motion for that task parameter.

Given a reference motion for each task parameter, one policy still has to follow the reference motion of the current task parameter without observing the task parameter.
This can be achieved by decoupling the observation from the reward, so that the reward tracks a reference motion that the policy does not observe.
Similar decoupling is common in motion imitation, where the reward contains more information about the reference than the policy observes~\cite{peng2018deepmimic, peng2021amp, xu2025facet}.
SoftMimic~\cite{margolis2025softmimic} uses this idea to learn compliance.
It generates compliant responses to external forces from one motion clip with an inverse kinematics model, and rewards the policy for matching these responses while the policy observes only the nominal motion.
Since the policy does not observe which response the reward expects, it learns to estimate the force from its state and to produce the matching response.
This decoupling works when the reward scores the correct response to the unobserved quantity, and when this quantity affects the state that the policy observes.
In box pushing, the reference motions generated for each task parameter give the target posture, and the response of the box to the push depends on the task parameter, as the force-adaptive policies show~\cite{zhang2025falcon, hu2026vofa, xu2025facet, zhi2025learning, portela2024learning}.
Therefore, the posture can be adapted to the task parameter without the task parameter as an input.

This paper presents PushMimic, a framework (Fig.~\ref{fig:teaser}) that augments the single level-ground nominal reference motion into an augmented motion dataset with one dynamically feasible reference motion for each task parameter (Section~\ref{sec:augmentation}), and distills this dataset into a single motion tracking policy in one stage of training, where the policy observes only the nominal reference motion (Section~\ref{sec:learning}).
In simulation on a Unitree G1 over a range of box masses and ramp inclinations, the policy adapts its posture to the box mass and the ramp inclination without observing them.
Compared with a policy that tracks only the nominal reference motion, it succeeds more often as the box gets heavier and the ramp steeper, and lowers the cost of push of the arms.
On the physical G1, the policy pushes a box continuously on an inclined treadmill and leans further forward as weights are added to the box during the push (Fig.~\ref{fig:hw_mass}).

The main contributions of this paper are:
\begin{itemize}
    \item A framework for adaptive humanoid box pushing from a single kinematic reference motion via dynamically feasible reference augmentation and one-stage policy distillation.
    \item Extensive simulation experiments on a Unitree G1 over box masses from $5$ to $15~\mathrm{kg}$ and ramp inclinations up to $15^\circ$, in which the proposed method raises the success rate from $57.8\%$ to $88.8\%$ compared with the vanilla baseline and lowers the cost of push of the arms by $22\%$.
    \item A sim-to-real transfer of the policy to the physical G1, which adapts its torso pitch during continuous pushing on an inclined treadmill as the box mass grows from $10$ to $19~\mathrm{kg}$ and as the ramp inclination rises from $0^\circ$ to $12.6^\circ$.
\end{itemize}

\section{Dynamic Motion Augmentation}\label{sec:augmentation}

This section presents how the nominal reference motion is augmented into a motion dataset with one reference motion for each task parameter. The augmentation uses sampling-based trajectory optimization (SBTO)~\cite{dhedin2026dynaretarget}.

\subsection{Problem Definition}\label{sec:problem}

This paper considers a humanoid robot pushing a large box up a ramp. The task parameter is $\theta=(m,\alpha)$, where $m$ is the box mass and $\alpha$ is the ramp inclination. Starting from the nominal reference motion recorded on level ground, the augmentation generates one dynamically feasible reference motion for each task parameter on a grid $\Theta$ of box masses and ramp inclinations.

\begin{figure}[htb]
    \centering
    \includegraphics[width=0.8\columnwidth]{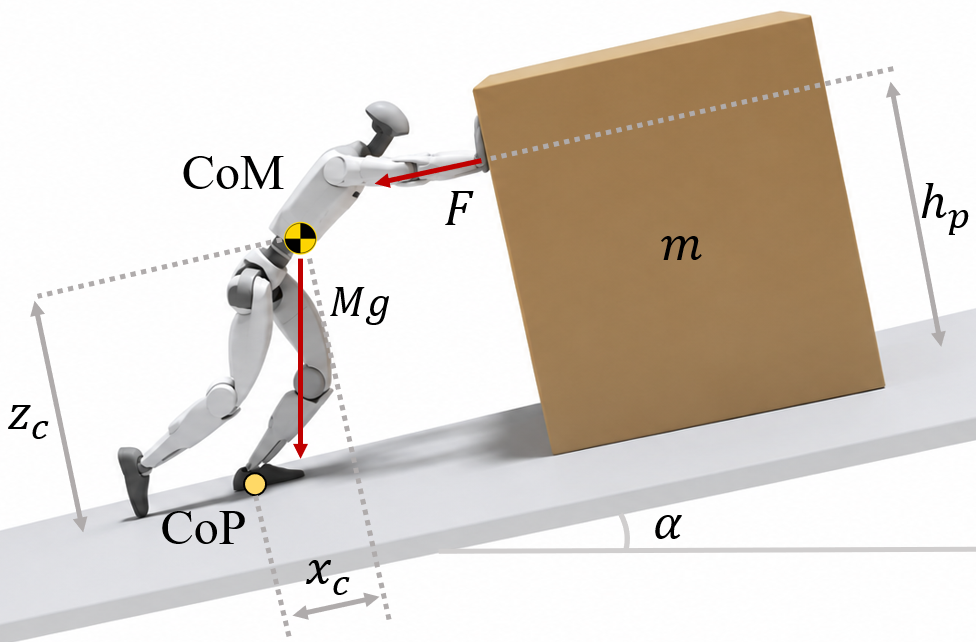}
    \caption{The simplified planar quasi-static model of the box pushing task.}
    \label{fig:quasi_static_model}
    \vspace{-1em}
\end{figure}

\subsection{The Simplified Model}\label{sec:aug_principle}

A planar quasi-static model gives intuition on how the pushing posture should change with the task parameter (Fig.~\ref{fig:quasi_static_model}).
When the robot pushes the box uphill, the push force $F = m g (\sin\alpha + \mu\cos\alpha)$ acts back on the hands and tends to tip the robot downhill.
Balancing the moments about the center of pressure (CoP) of the feet gives the lean angle $\psi$ of the line from the CoP to the robot center of mass (CoM), measured from the ramp normal,
\begin{equation}\label{eq:lean}
    \tan\psi = \frac{x_c}{z_c} = \tan\alpha + \frac{m}{M} \frac{h_p}{z_c} \left(\tan\alpha + \mu\right),
\end{equation}
where $g$ is the gravitational acceleration, $\mu$ is the friction coefficient between the box and the ramp, $M$ is the robot mass, $h_p$ is the height of the hands, and $x_c$ and $z_c$ are the distances from the CoP to the CoM along and normal to the ramp.
The first term is the lean of a robot that stands upright with respect to gravity, and the second term is the extra lean needed to resist the push, which grows with the box mass and the ramp inclination.
Since the CoP has to stay under the feet, the robot leans by pitching its torso forward and moving its feet back down the ramp.
The model assumes a fixed push height $h_p$, so the augmentation tracks the hand position in the box frame, which also prevents degenerate contacts such as pushing with the forearms.

\subsection{SBTO as Data Generator}

Generating a dynamically feasible reference motion for each task parameter requires optimizing over the full-order dynamics with intermittent contacts at the feet and the hands.
SBTO~\cite{dhedin2026dynaretarget} does not model these contacts explicitly, since it evaluates the cost on rollouts of a physics simulator, which resolves the contacts.
The control sequence is parameterized by a sparse set of knots of joint position targets, which a proportional-derivative (PD) controller tracks between knots.
In each iteration, a batch of knot sequences is sampled from a Gaussian distribution and rolled out in parallel in the simulator under the task parameter $\theta$.
Each rollout is scored with a cost, and the distribution is refit to the lowest-cost samples, following the cross-entropy method.
The optimization horizon is extended step by step, so that the earlier part of the motion converges before the later knots are optimized.

\subsection{SBTO Cost Design}\label{sec:aug_cost}

The cost $J$ of a rollout combines motion tracking, object tracking, and regularization terms,
\begin{equation}\label{eq:J}
J(\xi) =
J_{\mathrm{motion}}(\xi,\xi_0)
+ J_{\mathrm{object}}(\xi,\xi_0)
+ J_{\mathrm{reg}}(\xi),
\end{equation}
where $\xi$ is the trajectory of the robot and the box in the rollout, $\xi_0$ is the nominal reference motion, $J_{\mathrm{motion}}$ tracks the robot motion of $\xi_0$, $J_{\mathrm{object}}$ tracks the box motion of $\xi_0$, and $J_{\mathrm{reg}}$ penalizes undesirable contacts and constraint violations. All terms and weights are listed in Table~\ref{tab:cost}.

The motion tracking cost has two terms that let the posture adapt to the task parameter, as suggested by the simplified model in (\ref{eq:lean}),
\begin{equation}\label{eq:J_motion}
J_{\mathrm{motion}}
=
J_{\mathrm{track}}
+ \sum_k
\left(
c_{\mathrm{torso},k}
+
c_{\mathrm{foot},k}
\right),
\end{equation}
where $k$ indexes the frames of the rollout, and $J_{\mathrm{track}}$ tracks the joint positions, the base pose, and the hand position in the box frame.
The torso term $c_{\mathrm{torso},k}$ encourages a forward torso pitch, and the foot term $c_{\mathrm{foot},k}$ lets the feet move back within a dead zone.
Both terms keep the nominal reference motion as a prior while allowing these posture changes, as illustrated on the left of Fig.~\ref{fig:teaser}.

\subsubsection{Torso Pitch Prior}
The torso term tracks the nominal torso orientation pitched forward by an offset,
\begin{equation}\label{eq:lean_cost}
    c_{\mathrm{torso},k} = w_{\mathrm{t}} \left\| \vec{n}_k - \mat{R}(\phi)\, \vec{n}_{0,k} \right\|^2,
\end{equation}
where $\vec{n}_k$ is the vertical axis of the torso link at frame $k$ of the rollout, $\vec{n}_{0,k}$ is the same axis in the nominal reference motion, $\mat{R}(\phi)$ is the rotation about the lateral axis of the robot by the pitch offset $\phi$, and $w_{\mathrm{t}}$ is the weight.
The pitch offset is
\begin{equation}\label{eq:pitch_prior}
\phi(m,\alpha)=\alpha[\lambda(m-m_0)+1],
\end{equation}
where $m_0$ is the nominal box mass and $\lambda$ is a coefficient that controls the mass dependence.
Like (\ref{eq:lean}), the offset grows with the ramp inclination, and it grows faster for a heavier box.
The offset keeps only the inclination terms of (\ref{eq:lean}), and $\lambda$ is tuned.
The effect of the box mass on level ground is left to the simulated dynamics.

\subsubsection{Foot Placement Dead Zone}
The simplified model also suggests that the feet move back down the ramp as the inclination grows.
With a plain foot tracking cost, the feet stay at the level-ground footprints of the nominal reference motion, the robot cannot lean enough, and the box stalls on steep ramps.
A lower foot tracking weight frees the feet, but it also removes the nominal foot placement as a prior.
The foot term therefore uses a dead zone,
\begin{equation}\label{eq:deadzone}
    c_{\mathrm{foot},k} = w_{\mathrm{f}} \left\| \max\left(|\vec{e}_{\mathrm{f}}| - \bm{\delta}_{\mathrm{f}}, \zeros\right) \right\|^2,
\end{equation}
where $\vec{e}_{\mathrm{f}}$ is the foot position error relative to the nominal reference motion, $\bm{\delta}_{\mathrm{f}}$ is a per-axis dead-zone width, and $w_\mathrm{f}$ is the weight.
Within the dead zone, the optimizer places the feet freely according to the dynamics, and larger deviations are still penalized quadratically.
The dead zone is wider along the ramp, where the feet are expected to move back, and narrower in the lateral and vertical directions.

\section{Reference Distilled Motion Tracking}\label{sec:learning}

The augmented motion dataset gives the pushing posture for each task parameter, but the task parameter is unknown at deployment.
However, the task parameters are physical properties of the task, such as the box mass and the ramp inclination, so they change how the box and the robot respond to the push.
A policy can therefore infer the right posture from this response.
This section distills the dataset into a single policy that observes only the nominal reference motion and adapts its posture through this interaction.
The learning problem is formulated as a Markov decision process (MDP) in which the observation and the reward use different reference motions.

\subsection{MDP Design}\label{sec:il_mdp}

The MDP uses two reference motions, the nominal reference motion that the policy observes and the augmented reference motion that the policy is rewarded to track.
Training runs in parallel environments, and each environment holds one task parameter $\theta \in \Theta$ with its augmented reference motion $\hat{\xi}_\theta$.
The nominal reference motion $\xi_0$ is the same in every environment.
Both reference motions are sequences of frames at the control rate.
An episode starts at a frame $k_0$ and advances one frame per step, so that at step $k$ the nominal frame $\xi_{0,k}$ and the augmented frame $\hat{\xi}_{\theta,k}$ describe the same moment of the push.
At each step, the policy maps its observation $\ot_k$, built from the state $\st_k$ of the robot and the box, to the action $\at_k$, which is the joint position targets of the PD controller.

\subsubsection{Observations}
The observation is composed of three groups, $\ot_k = \big(\ot^{\mathrm{r}}_k, \ot^{\mathrm{b}}_k, \ot^{\mathrm{d}}_k\big)$,
    where $\ot^{\mathrm{r}}_k$ is the proprioception of the robot, $\ot^{\mathrm{b}}_k$ is the state of the box, and $\ot^{\mathrm{d}}_k$ is the nominal frame.
The proprioception contains the joint positions and velocities, the linear and angular velocity of the base, the gravity direction in the base frame, and the previous action.
The box state contains the pose and the twist of the box expressed in the robot torso frame.
The nominal frame contains the joint positions and velocities of $\xi_{0,k}$ and the pose of its torso relative to the torso of the robot.
The task parameter is not observed directly.
It affects the observation through the response to the push, for example, a heavier box moves more slowly under the same push, and a steeper ramp also tilts the gravity direction in the base frame.

\subsubsection{Rewards}
The reward is a sum of tracking terms and regularization terms,
\begin{equation}\label{eq:reward}
    r_k = \sum_{i} w_i \exp\!\left(-\frac{\|\vec{e}_{i,k}\|^2}{\sigma_i^2}\right) - \sum_{j} w_j\, c_{j,k},
\end{equation}
where $\vec{e}_{i,k}$ is the error of the $i$-th tracked quantity against the augmented frame $\hat{\xi}_{\theta,k}$, $\sigma_i$ is the width of its kernel, $c_{j,k}$ is the $j$-th regularization term, and $w_i$ and $w_j$ are the weights.
The tracked quantities are the pose of the torso, the poses and velocities of the tracked links relative to the torso, and the position, orientation, linear velocity and angular velocity of the box.
The box terms encode the task, which is to move the box along the reference box trajectory.
The regularization terms penalize the action rate and its second difference, joint limit violations, ankle joint position targets beyond the point where they saturate the ankle torque, deviations of the ankle roll targets from their default, and contacts of links other than the feet and the hands.
The two ankle terms keep the ankle targets away from torque saturation and from the ankle roll limit.

\subsubsection{Terminations and Resets}
The terminations keep the robot close to the augmented reference motion and make it push with the hands only.
An episode terminates when the torso height or tilt, or the box position or orientation, deviates from the augmented frame beyond a threshold.
Heights are measured normal to the ramp, so that moving along the ramp is not counted as a height error.
An episode also terminates when any link other than the hands contacts the box.
This contact rule stops the robot from pushing with a knee or a shoulder, which the reward alone does not reliably prevent.
At reset, the robot and the box are set to the augmented frame $\hat{\xi}_{\theta,k_0}$, with small perturbations of the base pose, the base velocity, and the joint positions.

\subsection{Distillation via Decoupled Observation and Reward}\label{sec:il_mechanism}
The key idea of the distillation is to reward the policy for a motion that it does not observe.
In an episode, the observation uses the nominal reference motion, while the reward and the initial state use the augmented reference motion,
\begin{equation}\label{eq:binding}
\begin{aligned}
    \ot_k &= h\big(\st_k, \xi_{0,k}\big), \\
    r_k &= r\big(\st_k, \at_k \mid \hat{\xi}_{\theta,k}\big), \\
    \st_{k_0} &\sim \hat{\xi}_{\theta,k_0},
\end{aligned}
\end{equation}
where $h$ builds the observation from the state and the nominal frame, and the reward (\ref{eq:reward}) is evaluated against the augmented frame.
To earn the reward, the policy has to infer the task parameter from how the box and the robot respond to the push, and produce the posture of the augmented reference motion.
Starting from the augmented reference motion puts the robot in the posture that the reward expects.

Since all augmented reference motions are variations of the same nominal reference motion at the same phase, one policy that observes the nominal frame can represent all of them.
The dataset is therefore distilled in one stage of training, without training one policy per reference motion and then distilling them.
The policy is trained with an asymmetric actor-critic algorithm.
The critic also observes the task parameter $\theta$ and the augmented frame $\hat{\xi}_{\theta,k}$ as privileged information.
Since the reward depends on both, this reduces the variance of the value estimate.

\section{Results}\label{sec:result}

This section presents simulation and hardware experiments that evaluate whether one level-ground reference motion is enough to push a box over a range of box masses and ramp inclinations.

\subsection{Experiment Setup}\label{sec:setup}

\subsubsection{Robot Hardware}
\begin{figure}[htb]
    \centering
    \includegraphics[width=\columnwidth]{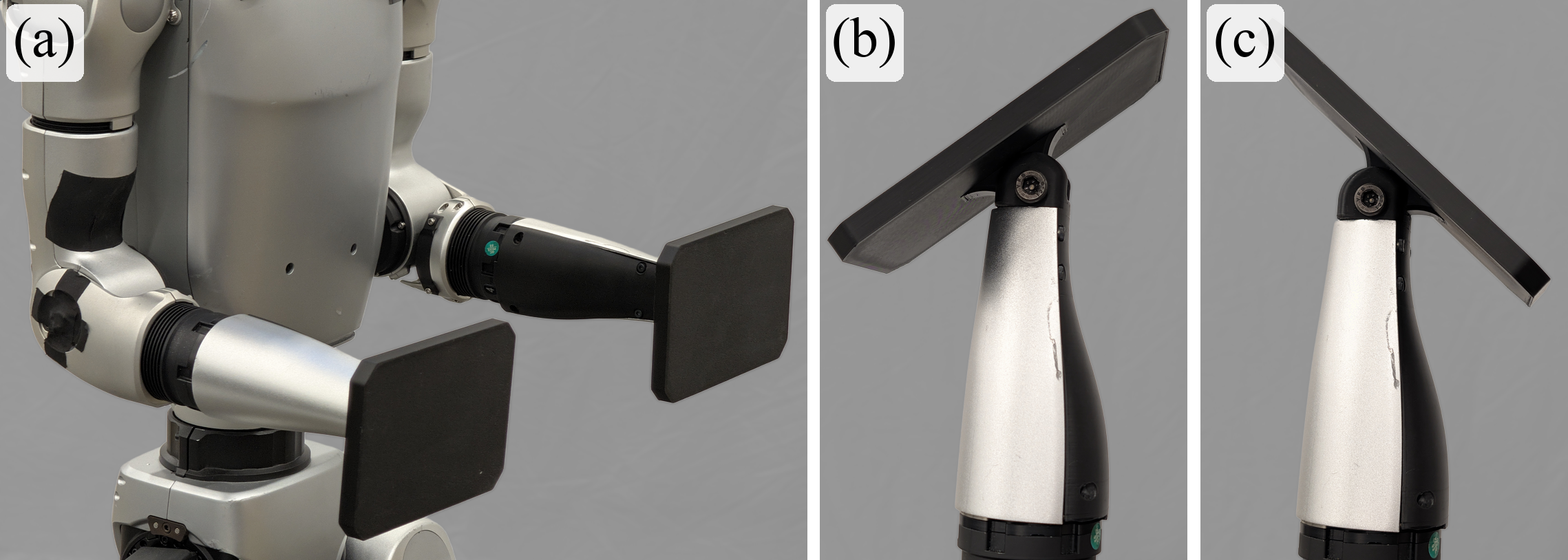}
    \caption{The flat hands with passively damped hinges that replace the stock hands of the G1 and enlarge the contact area on the box. (a) both hands mounted on the arms, and the hinge at its (b) upper and (c) lower limit.}
    \label{fig:flat_hand}
    \vspace{-1em}
\end{figure}

The framework is validated on a Unitree G1 humanoid robot with 23 actuated joints, in simulation and on hardware.
The stock hands are removed and replaced by a flat plate on a passive damped hinge, so that the plate keeps face contact with the box while the torso pitch changes (Fig.~\ref{fig:flat_hand}).

\subsubsection{Nominal Reference Motion}
The nominal reference motion is a human pushing motion generated by Kimodo~\cite{rempe2026kimodo} and retargeted to the G1 with GMR~\cite{araujo2025retargeting}.
The box trajectory follows the running maximum of the leading hand position along the push direction, so that the box never moves backward and stays in contact with the leading hand.
The motion pushes a $0.9 \times 0.9 \times 1.0~\mathrm{m}$ box with the nominal mass $m_0 = 10~\mathrm{kg}$ forward by $3~\mathrm{m}$ in $9~\mathrm{s}$ on level ground.

\begin{figure}[htb]
    \centering
    \includegraphics[width=\columnwidth]{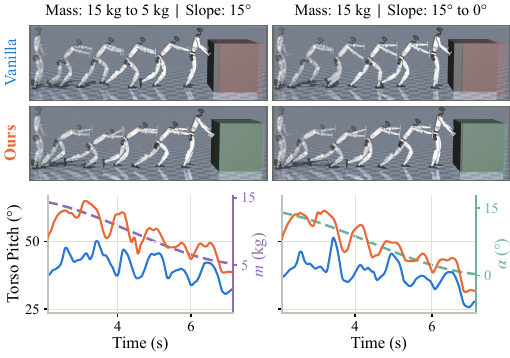}
    \caption{Online posture adaptation when the box mass or the ramp inclination changes during the push.
    The left column shows a push in which $m$ falls from $15$ to $5~\mathrm{kg}$ on a $15^\circ$ ramp.
    The right column shows a push in which $\alpha$ falls from $15^\circ$ to $0^\circ$ with a $15~\mathrm{kg}$ box.
    Top and middle rows: strobe renders of the vanilla baseline and of PushMimic, with the goal position of the box drawn in green when the push delivered the box and in red when it did not.
    Bottom row: torso pitch over time, with the task parameter as a dashed line on the right axis.}
    \label{fig:fig_varying_row}
    \vspace{-1em}
\end{figure}

\subsubsection{Augmented Motion Dataset Generation}
The task parameter grid $\Theta$ covers box masses from $5$ to $15~\mathrm{kg}$ in steps of $1~\mathrm{kg}$ and ramp inclinations from $0^\circ$ to $15^\circ$ in steps of $1^\circ$, which gives 176 augmented reference motions.
The box-to-ramp friction coefficient is $0.5$, and the training environments use the same friction coefficients as the augmentation.
Each reference motion is solved in about $9~\mathrm{min}$ on a desktop with an Intel i7-14700K CPU.
SBTO draws 1024 knot sequences per iteration and runs at most 50 iterations each time the horizon is extended.

\subsubsection{Motion Tracking Policy Training}
Policies are trained with proximal policy optimization (PPO)~\cite{schulman2017proximal} in Isaac Lab on an NVIDIA RTX 4090 GPU, with 4096 parallel environments for 10,000 iterations, which takes about $4~\mathrm{h}$ per run.
Three seeds are trained per variant.
The reward terms follow (\ref{eq:reward}), and the kernel widths and weights are listed in Table~\ref{tab:reward}. Domain randomization follows~\cite{liao2026beyondmimic}, with the external push on the robot disabled.

\subsubsection{Baseline}
The vanilla baseline is the standard motion tracking recipe of~\cite{liao2026beyondmimic} applied to the nominal reference motion without augmentation.
It observes the same nominal frame as the proposed method, so both policies receive the same information at deployment.
Its reward and initial state use the nominal reference motion rotated onto the ramp by $\mat{R}(\alpha)$, the rotation about the lateral axis by the ramp inclination, namely $r_k = r\big(\st_k, \at_k \mid \mat{R}(\alpha)\,\xi_{0,k}\big)$.
Everything else is shared with the proposed method, including the environments over the same grid of task parameters, the domain randomization, and the reward weights.

\subsubsection{Metrics}
Both methods are deployed sim-to-sim in MuJoCo and evaluated with the following metrics.

\textbf{Task completion.} A push succeeds when the box end error, the distance between the final box position and that of the nominal reference motion, is within $0.1~\mathrm{m}$.
The success over the grid is summarized by $F_{50}$, the push force $F$ of the simplified model (Section~\ref{sec:aug_principle}) at which a logistic fit of the success against $F$ crosses $50\%$.

\textbf{Adaptiveness.} The torso pitch averaged over the push is regressed linearly on the box mass and the ramp inclination, once for the rollouts and once for the augmented reference motions of the same task parameters.
The mass adaptiveness $k_m$ and the inclination adaptiveness $k_{\alpha}$ are the coefficients of the rollouts divided by those of the augmented reference motions.
A value of one means that the policy reproduces the posture change of the augmented motion dataset, a value of zero means that its posture ignores the task parameter, and a negative value means that its posture changes in the opposite direction.

\textbf{Cost of push.} Analogous to the cost of transport, the cost of push is defined as $C = E/(F d)$, where $E = \Delta t \sum_k \|\bm{\tau}_k\|^2$ is the thermal loss during the push, $\bm{\tau}_k$ is the vector of joint torques at step $k$, $\Delta t$ is the time step, $F$ is the push force, and $d$ is the distance that the box travels.
The cost of push of the arms $C_{\mathrm{arm}}$ counts only the 10 arm joints.

\begin{figure}[htb]
    \centering
    \includegraphics[width=\columnwidth]{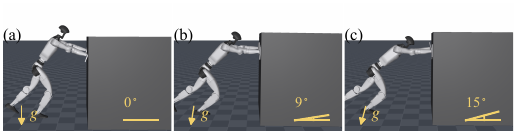}
    \caption{Augmented reference motions at the same frame of the push on ramps of (a) $0^\circ$, (b) $9^\circ$, and (c) $15^\circ$. The arrow $g$ shows the rotated gravity that simulates the ramp.}
    \label{fig:snapshot}
    \vspace{-1em}
\end{figure}

\subsection{Evaluation of Augmented Motion Dataset}\label{sec:result_dataset}

\begin{figure}[htb]
    \centering
    \includegraphics[width=\columnwidth]{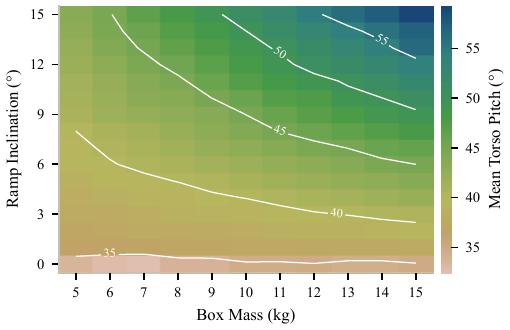}
    \caption{Mean torso pitch of the augmented reference motions during the push over the task parameter grid of box mass, from $5$ to $15~\mathrm{kg}$, and ramp inclination, from $0^\circ$ to $15^\circ$, one grid point per reference motion, with iso-pitch contours every $5^\circ$.}
    \label{fig:dataset_grid}
    \vspace{-1em}
\end{figure}

This subsection checks whether the augmentation produces the posture change suggested by the simplified model in (\ref{eq:lean}).
Fig.~\ref{fig:snapshot} shows the augmented reference motions at the same frame of the push on three ramps, where the ramp inclination is simulated as a rotation of gravity.
On a steeper ramp, the torso leans further forward and the feet stand further back.
Fig.~\ref{fig:dataset_grid} shows the torso pitch averaged over the push for every task parameter on the grid.
The pitch grows with both the ramp inclination and the box mass.
The iso-pitch contours run diagonally, so a heavier box on a gentler ramp needs the same posture as a lighter box on a steeper ramp.
On a $15^\circ$ ramp, the pitch rises from $43.6^\circ$ at $5~\mathrm{kg}$ to $59.3^\circ$ at $15~\mathrm{kg}$, which matches (\ref{eq:lean}), where a heavier box makes the lean grow faster with the inclination.
With the nominal $10~\mathrm{kg}$ box, the pitch grows from $34.7^\circ$ on level ground to $51.3^\circ$ on a $15^\circ$ ramp.
Over the same range, the feet move back, and the distance from the feet to the box grows by $0.19~\mathrm{m}$.

\subsection{Evaluation of Simulation Deployment}\label{sec:result_sim}
This subsection compares the proposed method with the vanilla baseline in simulation over the task parameter grid $\Theta$, in terms of task completion, adaptiveness, and cost of push.
The success rate and $F_{50}$ are computed over all rollouts, and the other metrics over the task parameters where each policy succeeds.
Table~\ref{tab:grid} lists the mean and the standard deviation over the three seeds.

\begin{table*}[htb]
    \caption{Policy evaluation over the task parameter grid in simulation.}
    \label{tab:grid}
    \centering
    \begin{tabular}{lcccccc}
        \toprule
         & Success Rate (\%) $\uparrow$ & $F_{50}$ ($\mathrm{N}$) $\uparrow$ & $k_{\alpha}$ $\uparrow$ & $k_m$ $\uparrow$ & $C_{\mathrm{arm}}$ ($\mathrm{N\,m\,s}$) $\downarrow$ & Box End Error ($\mathrm{m}$) $\downarrow$ \\
        \midrule
        Vanilla & $57.8 \pm 11.6$ & $66.2 \pm 8.5$ & $0.28 \pm 0.03$ & $-0.20 \pm 0.13$ & $3.62 \pm 0.07$ & $0.061 \pm 0.013$ \\
        \textbf{Ours} & $\bm{88.8 \pm 1.0}$ & $\bm{91.8 \pm 0.5}$ & $\bm{0.99 \pm 0.04}$ & $\bm{0.83 \pm 0.06}$ & $\bm{2.81 \pm 0.16}$ & $\bm{0.051 \pm 0.007}$ \\
        \bottomrule
    \end{tabular}
    \vspace{-1em}
\end{table*}

\begin{figure}[htb]
    \centering
    \includegraphics[width=\columnwidth]{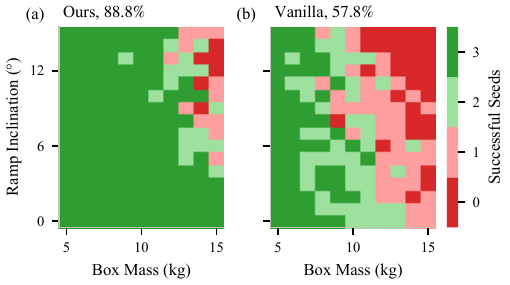}
    \caption{Success over the task parameter grid in simulation for (a) the proposed method and (b) the vanilla baseline, as the number of seeds whose push placed the box within the criterion, with the success rate over all rollouts in the panel label.}
    \label{fig:grid_success}
    \vspace{-1em}
\end{figure}

\begin{figure}[htb]
    \centering
    \includegraphics[width=\columnwidth]{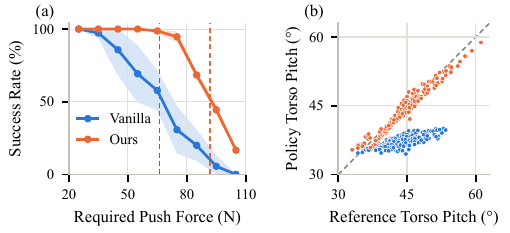}
    \caption{Pushing capacity and posture adaptation in simulation. (a) Success rate against the required push force $F = m g (\sin\alpha + \mu\cos\alpha)$ in bins of $10~\mathrm{N}$, as the mean over three seeds with the standard deviation shaded, and the force $F_{50}$ at which the success rate falls to $50\%$ (dashed). (b) Mean torso pitch of the policy during the push against that of the augmented reference motion, one point per task parameter where at least one seed succeeds, with the identity line dashed.}
    \label{fig:push_adapt}
\end{figure}

\subsubsection{Task Completion}\label{sec:result_task}
Fig.~\ref{fig:grid_success} shows the success over the grid.
The proposed method succeeds in $88.8\%$ of the rollouts, against $57.8\%$ for the baseline.
The baseline fails more often as the box gets heavier, while the proposed method succeeds in nearly every push up to $11~\mathrm{kg}$.
Fig.~\ref{fig:push_adapt}(a) combines the box mass and the ramp inclination into the push force $F$, and $F_{50}$ rises from $66~\mathrm{N}$ for the baseline to $92~\mathrm{N}$.
Where the policies succeed, both deliver the box well within the criterion.

\subsubsection{Adaptiveness}\label{sec:result_adapt}

Fig.~\ref{fig:fig_varying_row} shows the online posture adaptation within a single push, where the box mass or the ramp inclination changes while the box moves.
The torso pitch of the proposed method follows the change, and both pushes deliver the box.
The baseline keeps one posture and fails both pushes.
In Table~\ref{tab:grid}, the inclination adaptiveness is $0.99$ for the proposed method against $0.28$ for the baseline, and the mass adaptiveness is $0.83$ against $-0.20$.
The proposed method therefore reproduces almost all of the posture change with the inclination and most of that with the box mass, while the baseline follows about a quarter of the change with the inclination and leans slightly less as the box gets heavier.
Fig.~\ref{fig:push_adapt}(b) shows the same for each task parameter, where the torso pitch of the proposed method follows that of the augmented reference motion, while that of the baseline barely changes.

\subsubsection{Cost of Push}\label{sec:result_effort}
Where the policies succeed, the proposed method lowers the cost of push of the arms $C_{\mathrm{arm}}$ by $22\%$ compared with the baseline.
The two methods push the box with nearly the same hand force, and the advantage of the proposed method grows where its torso pitch is closer to the augmented reference motion.
That is because the leaned torso, rather than the arms, supplies the additional push.

\subsection{Hardware Experiment}\label{sec:result_hw}
This subsection tests whether the policy transfers to the physical G1 and adapts its posture when the box mass or the ramp inclination changes during a push.
The experiments use a treadmill with an adjustable inclination as an endless ramp, so that the task parameter can be changed while the robot keeps pushing.
For continuous pushing, the nominal reference motion is divided into a start segment, one pushing stride of $1.52~\mathrm{s}$, and a stop segment, and the stride is repeated, so that the reference box advances $0.94~\mathrm{m}$ per stride.
The belt runs at $0.6~\mathrm{m/s}$, which matches this pace, so the robot and the box stay in place on the treadmill.
The box of $0.8 \times 0.8 \times 1.0~\mathrm{m}$ weighs $10~\mathrm{kg}$, and weights are placed on top of it by hand during the push.
A motion capture system tracks markers on the box and on the torso of the robot.
The box state in the robot torso frame is computed from the two tracked poses, and the joint and base states come from the onboard sensors.
The policy runs at $50~\mathrm{Hz}$ and sends joint position targets to the PD controllers, as in simulation.

The torso pitch is measured from the belt normal and averaged over a moving window of one stride to remove the oscillation of the gait.
For comparison, the same policy is run in MuJoCo under the same box mass and ramp inclination over time, without observation noise or action delay, with the box of the nominal reference motion.

\begin{figure}[htb]
    \centering
    \includegraphics[width=\columnwidth]{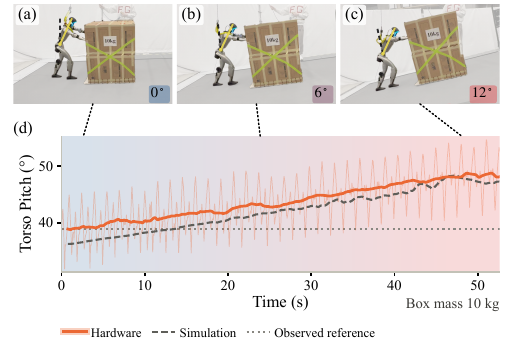}
    \caption{Online posture adaptation of the G1 when the ramp inclination rises during continuous pushing of a $10~\mathrm{kg}$ box. (a) to (c) Snapshots at inclinations of $0^\circ$, $6^\circ$, and $12^\circ$, with the guide lines of Fig.~\ref{fig:hw_mass}. (d) Torso pitch and its raw measurement (faint curve) as in Fig.~\ref{fig:hw_mass}(d), with the background colored by the inclination from blue at $0^\circ$ to red at $12.6^\circ$.}
    \label{fig:hw_incline}
    \vspace{-1em}
\end{figure}

\subsubsection{Varying Box Mass}
On the treadmill inclined at $10^\circ$, weights of $5$ and $4~\mathrm{kg}$ are placed on the box one after another, which raises the box mass from $10$ to $15$ and $19~\mathrm{kg}$ (Fig.~\ref{fig:hw_mass}).
The torso pitch rises within one stride of each placement, from $46.8^\circ$ with the bare box to $48.4^\circ$, and $52.8^\circ$.
In simulation, the torso pitch rises from $44.6^\circ$ to $54.7^\circ$ over the same masses.

\subsubsection{Varying Ramp Inclination}
With the $10~\mathrm{kg}$ box, the treadmill is tilted from $0^\circ$ to $12.6^\circ$ at a nearly constant rate over $46~\mathrm{s}$ while the robot pushes (Fig.~\ref{fig:hw_incline}).
The torso pitch grows from $39.1^\circ$ on level ground to $48.3^\circ$ at $12.6^\circ$, which is $0.71^\circ$ per degree of inclination, against $0.83^\circ$ in simulation.

Over nine continuous runs, including the two in Fig.~\ref{fig:hw_mass} and Fig.~\ref{fig:hw_incline}, the robot pushes for $8.5~\mathrm{min}$, or $307~\mathrm{m}$ of belt, without a fall.

\section{Conclusions}\label{sec:conclusion}

This paper presents PushMimic, a framework for adaptive humanoid box pushing from a single level-ground kinematic reference motion.
It augments the nominal reference motion into an augmented motion dataset with one dynamically feasible reference motion for each task parameter, and distills this dataset into a single motion tracking policy that observes only the nominal reference motion.
In simulation over box masses from $5$ to $15~\mathrm{kg}$ and ramp inclinations up to $15^\circ$, the policy adapts its posture to the box mass and the ramp inclination without observing them, with an adaptiveness of $0.83$ and $0.99$, respectively.
Compared with the vanilla baseline, it raises the success rate from $57.8\%$ to $88.8\%$ and lowers the cost of push of the arms by $22\%$.
On the physical Unitree G1, the policy adapts its torso pitch during continuous pushing as the box mass grows from $10$ to $19~\mathrm{kg}$ and as the ramp inclination rises from $0^\circ$ to $12.6^\circ$.

Two limitations motivate future work.
First, the augmented motion dataset is generated on a grid, whose size grows exponentially with the number of task parameters.
Generating reference motions for sampled task parameters, instead of solving every point on a grid, would extend the framework to more task parameters, such as the box size and the push distance.
Second, the box state is measured by a motion capture system, which limits the deployment to an instrumented space.
Estimating the box state with onboard sensors would allow deployment on real loading ramps.

\addtolength{\textheight}{-0cm}   %

\appendix

\begin{table}[H]
    \caption{Cost weights of the reference augmentation.}
    \label{tab:cost}
    \centering
    \begin{tabular}{lc}
        \toprule
        Cost Term & Weight \\
        \midrule
        \multicolumn{2}{l}{\textbf{Motion Tracking}} \\
        Joint positions & $0.04$ to $0.24$ \\
        Base position and orientation & $5.0$ / $0.02$ \\
        Hand position in box frame & $15$ \\
        \midrule
        \multicolumn{2}{l}{\textbf{Posture Adaptation}} \\
        Torso pitch prior & $w_t = 20,\lambda=0.1$ \\
        Foot placement dead zone & $15$ \\
        \midrule
        \multicolumn{2}{l}{\textbf{Object Tracking}} \\
        Box position and orientation & $40$ / $20$ \\
        \midrule
        \multicolumn{2}{l}{\textbf{Regularization}} \\
        Torso roll & $5.0$ \\
        Foot orientation bounds & $15$ / $5$ \\
        Box-ground contact schedule & $1.0$ \\
        \bottomrule
    \end{tabular}
\end{table}

\begin{table}[H]
    \caption{Reward weights of the policy training.}
    \label{tab:reward}
    \centering
    \begin{tabular}{lcc}
        \toprule
        Reward Term & Kernel Width & Weight \\
        \midrule
        \multicolumn{3}{l}{\textbf{Motion Tracking}} \\
        Torso position and orientation & $0.20$ / $0.22$ & $1.5$ \\
        Link position and orientation & $0.18$ / $0.22$ & $2.0$ \\
        Link linear and angular velocity & $0.70$ / $1.80$ & $1.0$ \\
        \midrule
        \multicolumn{3}{l}{\textbf{Object Tracking}} \\
        Box position and orientation & $0.12$ / $0.12$ & $2.0$ / $1.0$ \\
        Box linear and angular velocity & $0.35$ / $0.90$ & $0.5$ \\
        \midrule
        \multicolumn{3}{l}{\textbf{Regularization}} \\
        Action rate & -- & $0.3$ \\
        Action second difference & -- & $0.1$ \\
        Joint limit & -- & $10$ \\
        Ankle target limit & -- & $10$ \\
        Ankle roll target & -- & $5$ \\
        Undesired contact & -- & $0.1$ \\
        \bottomrule
    \end{tabular}
\end{table}

\expandafter\let\expandafter\v\csname OT1\string\v\endcsname
\bibliographystyle{IEEEtran}
\bibliography{reference}

\end{document}